\documentclass[letterpaper, 10 pt, conference]{ieeeconf}  

\IEEEoverridecommandlockouts                              

\title{\LARGE \bf
ObjRetarget: An Object-Aware Motion Retargeting Framework with Anthropomorphic Arm Constraints and Polyhedral Hand Modeling}

\author{Yuanchuan Lai$^{1}$, Qing Gao$^{1,*}$, Ziyan Liang$^{1}$, Junjie Hu$^{2}$, Zhaojie Ju$^{3}$
\thanks{This work was supported in part by the the Shenzhen Science and Technology Program under Grant ZDCY20250901100201002, in part by the Guangdong Basic and Applied Basic Research Foundation under Grant 2025A1515011954, 2023A1515110074.}
\thanks{$^{1}$Yuanchuan Lai, Qing Gao, Ziyan Liang are with the School of Electronics and Communica-tion Engineering, Sun Yat-sen University, Shenzhen 518107, China.(email:laiych25@mail2.sysu.edu.cn, gaoqing2@mail.sysu.edu.cn)}%
\thanks{$^{2}$Junjie Hu is with the School of Artificial Intelligence, The Chinese University of Hong Kong, Shenzhen, Shenzhen 518172, China.(email: hujunjie@cuhk.edu.cn
)}%
\thanks{{$^{3}$Zhaojie Ju is with the School of Computing, University of Portsmouth, Portsmouth PO1 3HE, UK.(email: Zhaojie.Ju@port.ac.uk)}%
}
\thanks{$^{*}$Corresponding Author:Qing Gao, gaoqing2@mail.sysu.edu.cn.}%
}

\usepackage{soul} 
\usepackage{color, xcolor} 
\usepackage{cite}
\usepackage{graphicx}
\usepackage{booktabs}
\usepackage{multirow}
\usepackage{amsmath}
\usepackage{amssymb}   
\usepackage[colorlinks,bookmarksopen,bookmarksnumbered,citecolor=green, linkcolor=red, urlcolor=green]{hyperref}
\hypersetup{
    colorlinks=true, 
    urlcolor=magenta,   
}

\begin{document}

\maketitle
\pagestyle{empty}  
\thispagestyle{empty}
\pagestyle{empty}

\begin{abstract}
Learning robot dexterous manipulation from human manipulation videos requires reliably retargeting human intent to executable robot actions while maintaining stable hand–object contact, which remains a key challenge in embodied intelligence. Existing retargeting methods often ignore explicit contact modeling or rely on reinforcement learning, resulting in limited accuracy and generalization. To address this, we propose ObjRetarget, a human-to-robot motion retargeting framework for learning robot dexterous manipulation from human videos, which integrates anthropomorphic arm trajectory constraints with structured hand–object geometric modeling. For arm motion, reference trajectories extracted from human videos are used for initialization, followed by anthropomorphic constraints and redundancy-aware optimization to generate natural and accurate movements. For hand manipulation, ObjRetarget represents multi-finger contacts using polytope clusters and preserves contact structure through geometric invariants to improve stability. Experiments on real robots show that ObjRetarget improves manipulation success rates and contact stability across multiple dexterous tasks, and generalizes well to different demonstrations, object poses, and task settings. 
Project can be found at: \href{https://3469627147abc.github.io/ObjRetarget/}{https://github.io/ObjRetarget.}

\end{abstract}

\section{Introduction}
\begin{figure}
\includegraphics[width=\linewidth]{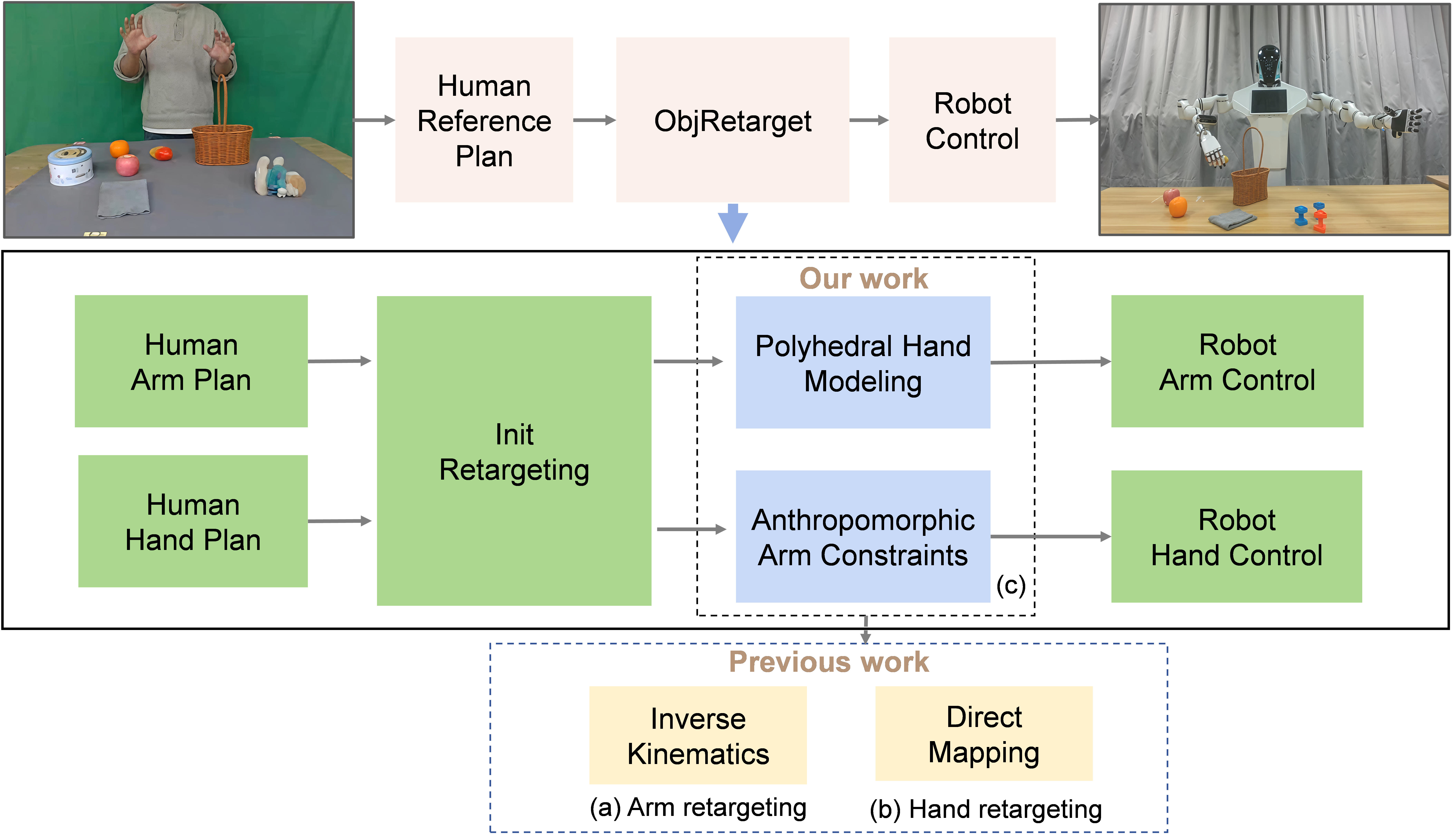}
\centering
\caption{Schematic of the motion retargeting framework. Part (a) shows arm retargeting via inverse kinematics, part (b) shows hand retargeting via direct mapping, and part (c) presents the proposed pipeline with polyhedral hand modeling and anthropomorphic arm constraints.} \label{figbegin}
\end{figure}

Learning robot dexterous manipulation from human manipulation videos requires reliably retargeting human intent into executable robot actions while maintaining stable hand–object contact throughout the manipulation process. This remains a central challenge in embodied intelligence and dexterous manipulation research, as successful manipulation depends not only on motion imitation but also on preserving physically meaningful contact relationships \cite{a1,a2,a3}. Existing motion retargeting approaches mainly focus on pose or joint mapping in free space, where the robot is encouraged to reproduce demonstrated kinematics \cite{a4,a5}, as illustrated in Fig. \ref{figbegin} (a). However, modeling object exchange, contact transitions, and sustained hand–object interaction is still limited, particularly in approaches that rely on direct hand mapping \cite{a6,a7} (Fig. \ref{figbegin} (b)). In real-world manipulation, robots must not only reproduce fine-grained finger motions but also coordinate large-scale arm movements, ensuring that grasping, transport, and object operation remain both physically stable and behaviorally natural. This requires reasoning about contact geometry, motion continuity, and task-dependent constraints simultaneously.

However, most existing methods treat the arm and hand as a unified system, overlooking their fundamental differences in motion scale, functional roles, and precision requirements \cite{a8}. The arm primarily produces large, global movements that determine object approach, spatial positioning, and overall motion trajectory, whereas the hand performs fine, contact-sensitive actions that directly govern grasp stability and manipulation success. Ignoring this distinction often leads to conflicts between global motion smoothness and local contact stability, making it difficult to simultaneously achieve robustness and naturalness in tasks such as object transfer, contact establishment, and contact maintenance. Meanwhile, many interaction-aware approaches rely on reinforcement learning for end-to-end optimization \cite{a9,a10}. Although such methods can implicitly learn contact behaviors, their generalization ability is often limited, and performance can degrade when object geometry, initial poses, or task conditions vary. These limitations suggest that stable and generalizable retargeting requires not only reproducing demonstration motions, but also explicitly modeling hand–arm motion differences and incorporating object interaction constraints into the retargeting process.

To address these challenges, we propose ObjRetarget, an object-interaction-oriented human-to-robot retargeting framework for dexterous manipulation, as illustrated in Fig. \ref{figbegin}. The pipeline first builds a human reference plan by extracting articulated joint trajectories and object point-cloud information from demonstration videos, followed by an initial retargeting stage that retargets human motion into a coarse robot trajectory for subsequent optimization. Building upon this initialization, ObjRetarget explicitly models the distinct motion characteristics of the arm and hand. For arm motion, we propose to refine the retargeted reference trajectories through anthropomorphic motion constraints and task-adaptive optimization, enabling the global motion to follow human movement patterns while adapting to variations in object initial poses, thereby improving motion naturalness and accuracy. For hand manipulation, we propose to construct polytope clusters in which each finger–object contact forms a local geometric unit, and geometric invariant constraints are applied to preserve contact structure and manipulation semantics, enhancing grasp stability and interaction reliability. Extensive real-robot experiments show that ObjRetarget significantly improves manipulation success rates and contact stability across multiple dexterous manipulation tasks, while maintaining natural and interpretable robot motions.

Our contributions are summarized as follows:
\begin{itemize}

\item We propose ObjRetarget, a hand–arm decoupled human-to-robot retargeting framework that explicitly separates global arm motion from contact-sensitive hand manipulation, enabling object-aware motion generation within a unified pipeline.

\item We develop a reference-trajectory-guided arm optimization strategy with anthropomorphic motion constraints, producing stable and human-consistent arm trajectories under varying object poses and task conditions.

\item We introduce a polytope-cluster hand modeling method that represents multi-finger contacts as local geometric units and preserves hand–object contact structure through geometric constraints.

\end{itemize}

\section{Related Work}
\subsection{Object-Aware Dexterous Manipulation}
Object-aware dexterous manipulation is challenging due to multi-contact constraints and complex hand–object geometry. Reinforcement learning and model predictive control are widely used to learn interaction policies \cite{a26,a27}, but reinforcement learning typically requires extensive training processes with large-scale interaction rollouts and careful reward design, and its performance can be sensitive to variations in object geometry and initial conditions \cite{a28,b1}. Other approaches incorporate explicit contact models or tactile sensing to improve stability \cite{a29,b2}, yet they often depend on accurate contact modeling or specialized tactile hardware, limiting scalability and cross-object generalization. As a result, achieving stable and generalizable dexterous manipulation without heavy training procedures or precise object modeling remains difficult.

To overcome these limitations, ObjRetarget leverages human manipulation priors from videos rather than relying on autonomous policy learning. It models hand–object interaction using polytope clusters with geometric invariants and guides arm motion with reference trajectories from demonstrations. By avoiding large-scale training procedures and explicit object models, this structured, human-informed formulation enables stable interaction and improved generalization across diverse objects and task scenarios.

\subsection{Human-to-Robot Motion Retargeting}
Human-to-robot motion retargeting transfers human demonstrations to robot platforms with different kinematics and degrees of freedom for high-fidelity motion reproduction. Traditional methods rely on analytical kinematics, optimization, or constraint-based planning, typically using joint-space mapping to adapt arm and hand poses \cite{a11,a12,a13}, while more recent works employ deep learning or latent representations to reduce kinematic discrepancies and improve retargeting accuracy \cite{a14,a15,b4}. These approaches perform well in free-space imitation and trajectory tracking.

However, most methods assume object-free or weak-contact settings and focus on geometric consistency or joint error minimization, without explicitly modeling hand–object interactions. In addition, many approaches treat the arm and hand as a unified system \cite{a16,a17}, overlooking their distinct roles: the arm controls large-scale motion, while the hand requires precise, contact-sensitive manipulation. This simplification can reduce coordination between global motion and local interaction, degrading performance in complex tasks.

ObjRetarget addresses these limitations by separating arm-level global motion from hand-level fine manipulation. Arm motion is guided by reference trajectories and refined with task-related constraints, producing physically executable motions that preserve human characteristics and support stable object interaction. At the same time, the retargeting process explicitly accounts for hand–object contact, ensuring stable and precise multi-finger manipulation during interaction.

\subsection{Learning from Human Demonstrations}
Human demonstration videos provide rich priors for robot dexterous manipulation, enabling object-aware motion retargeting from humans to robots. Traditional imitation-learning methods map human videos or motion capture data to robot control policies through trajectory replay, behavior cloning, or end-to-end visuomotor learning \cite{a18,a19,a20,a22}. While effective for skill acquisition, these approaches typically focus on autonomous policy learning and often ignore fine-grained hand–object interactions, limiting precise motion mapping and stable grasp execution.

Single-video retargeting methods, such as OKAMI \cite{a32} and ORION \cite{a33}, leverage human demonstrations to generate executable robot motions. OKAMI imitates humanoid motion for skill learning but struggles with multi-stage operations and continuous contact control due to the lack of explicit hand–object modeling. ORION uses open-world object graphs to infer object-centric actions, improving generalization across objects and layouts, but does not model fine-grained hand–object contacts. These examples highlight the potential of video-driven manipulation while showing the need for structured, object-aware retargeting. ObjRetarget addresses this by modeling hand–object contacts with polytope clusters and guiding arm motion via human reference trajectories, enabling contact-consistent, anthropomorphic, and generalizable motion suitable for downstream skill learning.

\begin{figure*}
\includegraphics[width=\linewidth]{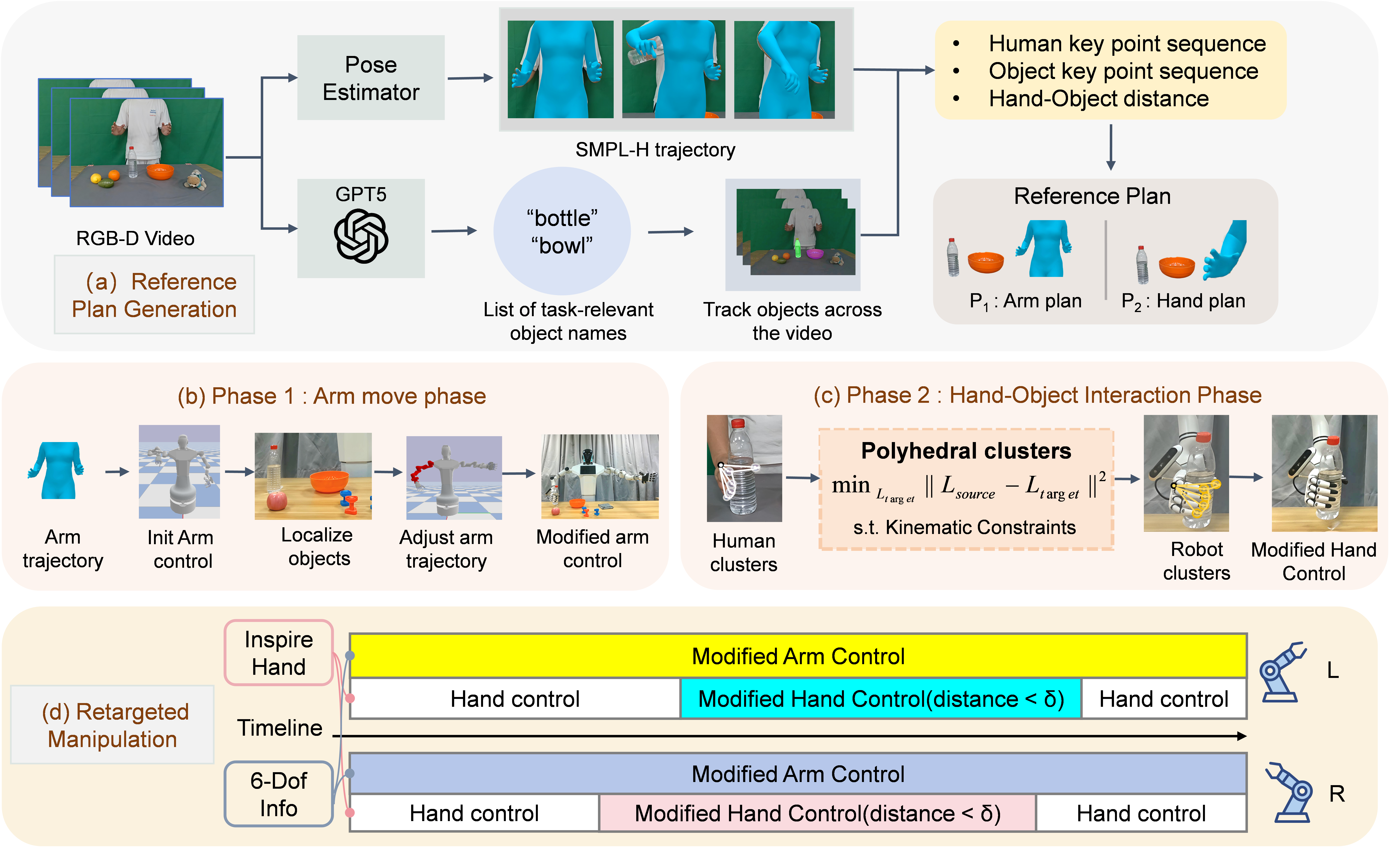}
\centering
\caption{Overview of the object-aware motion retargeting framework. Part (a) illustrates the extraction of human 3D body and hand poses from an RGB-D video, along with the detection and tracking of task-relevant objects to generate a reference plan. Part (b) presents arm trajectories refined based on object poses and anthropomorphic constraints. Part (c) shows hand motions optimized at contact points using polyhedral clusters. Part (d) depicts a unified temporal scheduler that synchronizes arm and hand motions for coordinated dual-arm execution.} \label{fig1}
\end{figure*}
\vspace{-8pt}

\section{ObjRetarget}
\subsection{Object-Aware Motion Retargeting Framework}
We propose ObjRetarget, an object-interaction-oriented motion retargeting framework that learns robot manipulation from human operation videos. As illustrated in Fig. \ref{fig1} (a), given an RGB-D manipulation video, human body motion is first reconstructed using the SLAHMR \cite{a30} pose estimator, an iterative optimization-based algorithm for recovering human motion sequences. The manipulated objects are then identified using a vision–language model, followed by object pose tracking to estimate their spatial trajectories over time. These components are integrated to obtain synchronized sequences of human joint poses, object point-cloud poses, and human–object interaction distances. Contact events are detected by thresholding the interaction distance, enabling the system to distinguish between free motion and physical interaction phases.
\vspace*{-0.3mm}

At the arm level (Fig. \ref{fig1} (b)), ObjRetarget generates global motion through a two-stage strategy. Human joint pose sequences are first used to produce an initialization retargeting trajectory, providing a feasible and structurally consistent motion prior. This trajectory is then refined by incorporating object pose information together with anthropomorphic motion constraints, yielding an executable arm trajectory that remains consistent with the demonstration while adapting to specific task configurations. At the hand level (Fig. \ref{fig1} (c)), ObjRetarget adopts a contact-aware staged retargeting mechanism. During the non-contact phase, the system follows the initialized hand motion obtained from retargeting. Once contact is detected, polytope-cluster-based geometric consistency optimization is activated to preserve local contact structures and manipulation semantics. 

Finally, Fig. \ref{fig1} (d) shows the unified temporal scheduler responsible for coordinating arm and hand trajectories. By synchronizing global arm motion with fine-grained hand manipulation, ObjRetarget ensures coherent dual-arm execution. Leveraging this hand–arm decoupled formulation, the framework achieves natural, stable, and generalizable dexterous manipulation in complex object-interaction tasks.

\subsection{Anthropomorphic Arm Constraints}
To obtain stable, natural, and generalizable robot execution trajectories in complex object-interaction tasks, ObjRetarget adopts a two-stage arm motion generation strategy. First, a reference trajectory is generated based on our previous retargeting work \cite{a31}, providing a feasible and structured initial motion. Then, the trajectory is refined using an optimization framework guided by anthropomorphic motion constraints and task objectives, producing motions that satisfy task requirements while maintaining anthropomorphic motion patterns. 

\begin{figure}
\includegraphics[width=\linewidth]{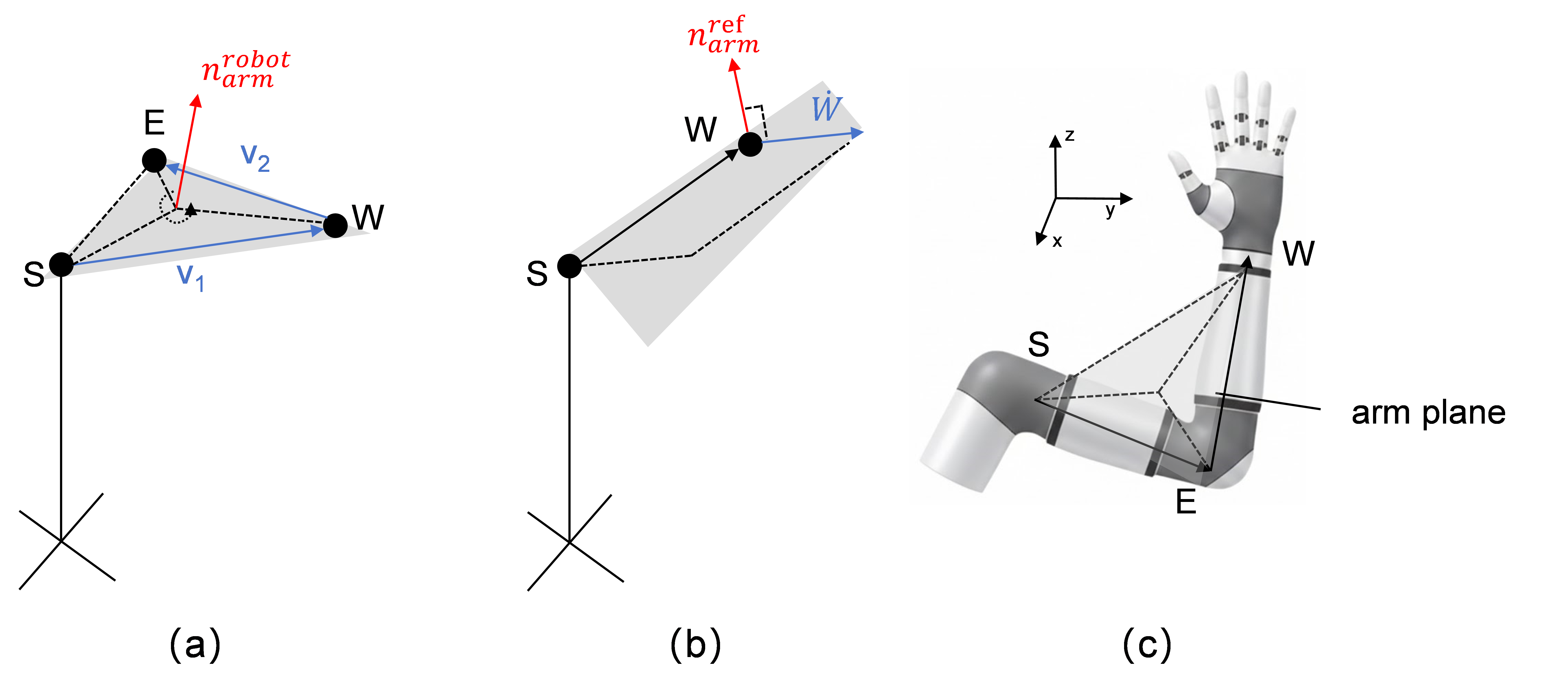}
\centering
\caption{Schematic of the task-adaptive arm-plane regularization. (a) The normalized normal vector ${n}_{{arm}}^{{robot}}$ defined by the shoulder-elbow-wrist triangle. (b) The reference normal ${n}_{{arm}}^{{ref}}$ constructed from the wrist motion direction. (c) The arm plane and its normal vector in the robot configuration.} \label{fig3}
\end{figure}

\textbf{\textit{Motion Initialization via Motion Retargeting}}: Although the initialization trajectory may not be the final execution result, it provides a structurally valid and near-optimal starting point, which greatly improves convergence stability in subsequent optimization.

Let the human motion sequence, consisting of $T$ frames, be represented as a set of skeleton graphs ${D}=\{G_k\}_{k=1}^{T}$, where each graph $G_k = (V_k, E_k, W_k)$ consists of joint nodes $V_k$, skeletal connectivity $E_k$, and dynamic attention weights $W_k$ at frame $k$. Each skeleton graph is embedded into a latent space via an encoder $f_\psi(\cdot)$, producing a compact representation $z_k = f_\psi(G_k)$, which is then decoded into the robot joint configuration $\theta_k = f_\phi(z_k) \in {R}^n$, where $n$ is the number of robot joints and $\theta_k$ respects the joint limits.

The corresponding robot end-effector pose $S_k$ is obtained via forward kinematics, $S_k = FK(\theta_k)$, where $S_k$ encodes both position and orientation of the end-effector. The retargeting process is optimized by minimizing a reference loss:
\begin{equation}
L_{{ref}}(S_k, S_k^h) = L_{{pos}} + \lambda_{{ori}} L_{{ori}} + \lambda_{{tip}} L_{{tip}},
\end{equation}
which accounts for end-effector position error $L_{{pos}}$, orientation error $L_{{ori}}$, and fingertip consistency $L_{{tip}}$ relative to the human demonstration $S_k^h$, with $\lambda_{{ori}}$ and $\lambda_{{tip}}$ weighting the contributions. The optimization over encoder and decoder parameters $(\psi, \phi)$ is formulated as:

\begin{equation}
\min_{\psi,\phi} \sum_{k=1}^{T} L_{{ref}}(S_k, S_k^h),
\end{equation}
\begin{equation}
{s.t.} \quad \theta_{\min} \le \theta_k \le \theta_{\max}.
\end{equation}

The resulting robot trajectory $S = \{S_k\}_{k=1}^{T}$ serves as the macro-level arm reference, providing an effective initialization for subsequent geometric consistency optimization. 

\textbf{\textit{Task-Adaptive Arm-Plane Regularization}}:
In human motion, the plane formed by the shoulder, elbow, and wrist evolves smoothly over time, and its normal direction is closely related to motion direction and task semantics. In contrast, robots with redundant degrees of freedom often produce non-anthropomorphic elbow configurations when only end-effector constraints are enforced, including elbow flipping, jitter, or abnormal lifting.

Existing approaches typically promote anthropomorphic motion by directly matching the arm-plane normal of the human demonstration. However, this strategy effectively treats the arm plane as a rigid pose target, which restricts motion flexibility in constrained manipulation scenarios. To maintain anthropomorphic structure while allowing adaptive motion, the arm plane is instead formulated as a soft geometric constraint.

As shown in Fig. \ref{fig3}, let the robot shoulder, elbow, and wrist positions at time $t$ be ${p}_s(t)$, ${p}_e(t)$, ${p}_w(t)$. The instantaneous arm-plane normal is defined as:

\begin{equation}
{n}_{{arm}}^{{robot}}(t)=
\frac{({p}_e(t)-{p}_s(t)) \times ({p}_w(t)-{p}_e(t))}
{\|({p}_e(t)-{p}_s(t)) \times ({p}_w(t)-{p}_e(t))\|}.
\end{equation}

Instead of using the human demonstration directly, the reference bending direction is constructed adaptively from the current motion trend:
\begin{equation}
{n}_{{arm}}^{{ref}}(t)=
\frac{({p}_w(t)-{p}_s(t)) \times \dot{{p}}_w(t)}
{\|({p}_w(t)-{p}_s(t)) \times \dot{{p}}_w(t)\|},
\end{equation}
where $\dot{{p}}_w(t)$ denotes the wrist velocity direction. The resulting task-adaptive arm-plane loss is defined as:
\begin{equation}
{L}_{{plane}}(t)
= w(t)\left(1-
\mathbf{n}_{{arm}}^{{robot}}(t)\cdot
\mathbf{n}_{{arm}}^{{ref}}(t)\right),
\end{equation}

This adaptive weighting $w(t)$ emphasizes the prior during large, fast arm motions while automatically reducing it during fine contact manipulation, preventing interference with task accuracy.
The formulation penalizes only large deviations from human-consistent bending trends while preserving redundancy flexibility.

\textbf{\textit{Task-Oriented End-Effector Tracking Loss}}:  
To ensure task accuracy during trajectory refinement, ObjRetarget includes a standard end-effector pose tracking objective. Let the wrist pose at time $t$ be $x_w(t) = [p_w(t), R_w(t)]$,
where $p_w(t) \in {R}^3$ is the wrist position and $R_w(t) \in SO(3)$ is the wrist orientation represented as a rotation matrix. Similarly, the reference wrist pose is $
x_w^{\text{ref}}(t) = [p_w^{\text{ref}}(t), R_w^{\text{ref}}(t)]$, where $p_w^{\text{ref}}(t)$ and $R_w^{\text{ref}}(t)$ denote the target position and orientation at frame $t$. The task-oriented tracking loss is defined as:
\begin{equation}
L_{\text{task}}(t) = \| p_w(t) - p_w^{\text{ref}}(t) \|^2 + \lambda_R \| \phi(R_w^{\text{ref}}(t)^\top R_w(t)) \|^2,
\end{equation}
where $\phi(\cdot)$ maps the rotation error from $SO(3)$ to its Lie algebra representation, and $\lambda_R$ weights the orientation term relative to the position term. This loss focuses purely on task accuracy and does not impose any additional priors, ensuring generality across different scenarios.

\textbf{\textit{Overall Optimization Objective}}: Combining task constraints and anthropomorphic regularization, the arm trajectory optimization is formulated as:
\begin{equation}
\begin{aligned}
\min_{{q}(t)}
& \sum_{t}
\Big[
{L}_{\text{task}}(t)
+ \lambda_{{p}}{L}_{{plane}}(t)
+ \lambda_{{s}}
\|{q}(t)-{q}(t-1)\|^2
\Big],
\end{aligned}
\end{equation}
where ${q}(t)$ denotes the robot joint configuration at time $t$, and $\lambda_{p}$ and $\lambda_{s}$ control the anthropomorphic regularization and temporal smoothness terms, respectively.

\subsection{Polyhedral Hand Modeling}
\textbf{\textit{Hand–object contact representation}}: 
To accurately capture fine-grained geometric relationships between the human hand and objects during multi-finger manipulation, ObjRetarget introduces an object-aware hand–object contact representation. Given an RGB-D input video, the system first extracts hand keypoints and the object surface point cloud. The hand keypoints include the five fingertips and the palm center (six points in total) to represent the overall hand configuration, while the object surface is reconstructed from the depth point cloud.

Based on the spatial adjacency between hand keypoints and object points, the system performs finger-wise contact detection to determine which fingers are in effective contact with the object. For a contacting finger $f$, its local contact region ${C}_f$ on the object surface is extracted, and the centroid of this region is used as the representative contact point:

\begin{equation}
{c}_f = \frac{1}{|C_f|} \sum_{{o} \in C_f} {o},
\end{equation}
where $|C_f|$ denotes the number of points in the contact region, and ${o}$ represents points on the object surface. This design enables robust extraction of multi-finger contact information under complex interaction scenarios, providing reliable input for structured geometric modeling.

\textbf{\textit{Polytope cluster construction}}: After obtaining finger-wise contacts, ObjRetarget constructs polytope clusters to model local geometric relationships during multi-finger collaboration. For a contacting finger $f$, a local tetrahedral unit ${T}_f$ is defined by four vertices: the palm center, the fingertip keypoint of $f$, the fingertip of an adjacent finger, and the corresponding object contact point ${c}_f$. When multiple fingers contact simultaneously, the resulting tetrahedra form a polytope cluster representing the local hand–object geometry under coordinated manipulation. This formulation transforms complex multi-finger contact relationships into a set of stable, compact geometric units, providing interpretable and optimizable high-dimensional contact constraints.

The core idea is that human manipulation does not require exact reproduction of absolute poses but rather the preservation of stable local geometry and contact semantics. Consequently, polytope clusters serve as soft geometric constraints, providing a structured basis for subsequent geometric consistency optimization.

\textbf{\textit{Geometric-invariant optimization}}: Based on polytope-cluster modeling, ObjRetarget formulates hand motion retargeting as a nonlinear optimization problem under geometric invariants. The overall geometric consistency loss is defined as:
\begin{equation}
{L} = \sum_f w_f \cdot {L}(T_f),
\end{equation}
where $f$ indexes the fingers, ${w}_f$ is a weight coefficient, and 
${L}(T_f)$ is the geometric consistency loss for the $f$-th tetrahedron, composed of edge-length and relative-pose invariants:
\begin{equation}
{L}(T_f) = {L}_{{edge}}(T_f) + \lambda {L}_{{pose}}(T_f),
\end{equation}

The edge-length loss constrains the local tetrahedral geometry:
\begin{equation}
{L}_{{edge}}(T_f) = \sum_{(i,j) \in E_f} 
\left\| {p}_i^r - {p}_j^r - ({p}_i^h - {p}_j^h) \right\|^2,
\end{equation}
where $E_f$ denotes the edges of the tetrahedron, and ${p}^r$,${p}^h$  are vertex positions in the robot and human demonstration, respectively.

To distinguish cases where geometric structure is preserved but manipulation semantics differ, a relative-pose invariant is introduced. Using a local coordinate frame based on hand keypoints, the relative position of the object contact point is constrained:
\begin{equation}
{L}_{{pose}}(T_f) = 
\left\| {R}_r^\top ({c}_f^r - {p}_0^r) - 
{R}_h^\top ({c}_f^h - {p}_0^h) \right\|^2,
\end{equation}
where ${p}_0$ is the palm reference point, and ${R}_r,{R}_h\in SO(3)$ represent the local hand frames of the robot and human demonstration, respectively. This term removes global translation and rotation while preserving the contact direction and spatial distribution relative to the hand, effectively maintaining manipulation semantics.

Finally, hand motion retargeting is formulated as the following constrained optimization problem:
\begin{equation}
\begin{aligned}
q^{*}=\min_{{q}} & \sum_f w_f ( {L}(T_f({q})) ), 
\end{aligned}
\end{equation}

\begin{equation}
{s.t.}  {q}_{min} \le {q} \le {q}_{max},
\end{equation}
where ${q}_{min}$, ${q}_{max}$ are joint limits, and $q^{*}$ denotes the optimized joint configuration achieving minimal combined edge and pose loss.

\begin{figure*}
\includegraphics[width=\linewidth]{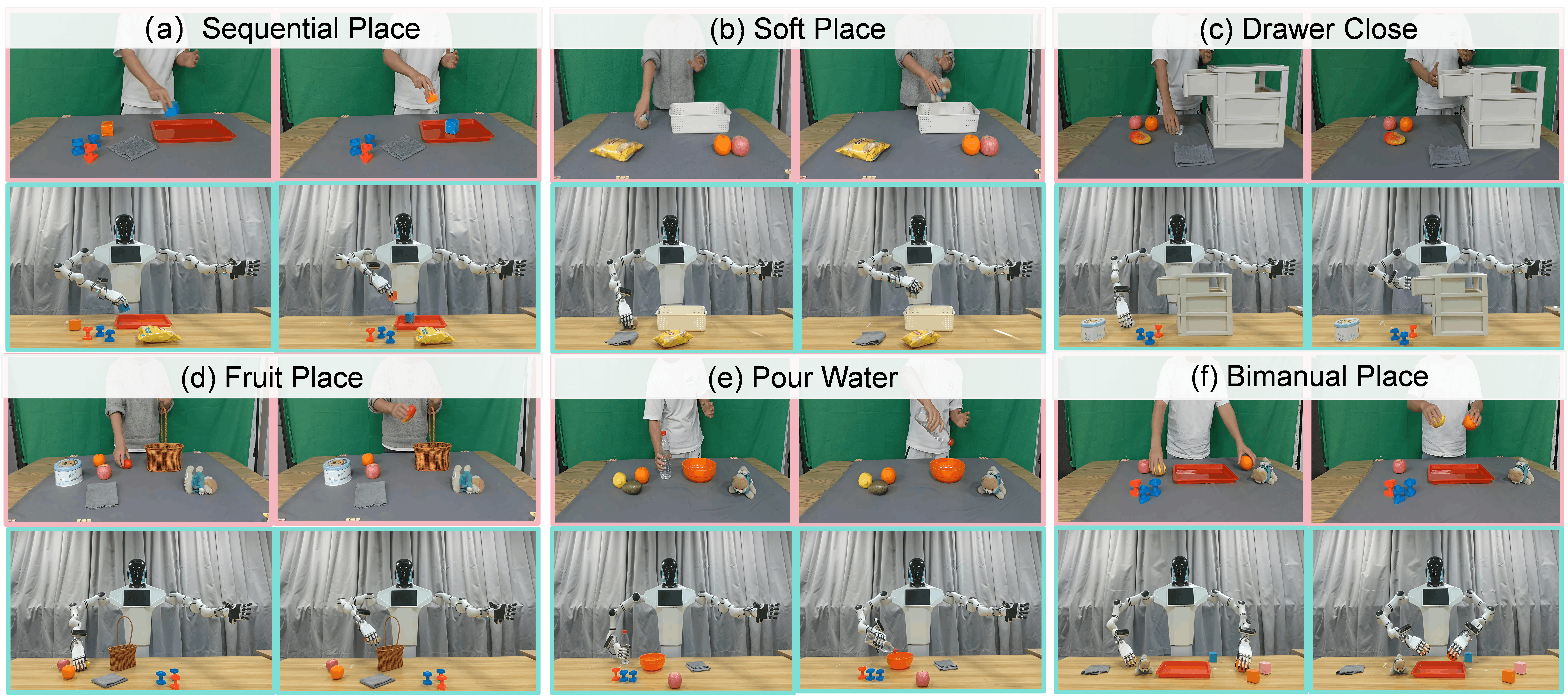}
\centering
\caption{Visualization of ObjRetarget on six real-world dexterous manipulation tasks. For each task, the top row shows the human demonstration and the bottom row shows the corresponding robot execution. From left to right: (a) Sequential Place: sequentially placing two cartons onto a tray, (b) Soft Place: grasping a plush toy and placing it into a basket, (c) Drawer Close: placing a medicine box into an open drawer and closing it, (d) Fruit Place:  transferring a mango from the table to a fruit basket, (e) Pour Water: picking up a bottle, pouring water into a container, and returning it, and (f) Bimanual Place: bimanual grasping of an apple and a lemon followed by synchronized placement.} \label{fig4}
\end{figure*}

\subsection{Synchronized Arm–Hand Execution}
After obtaining the optimized arm and hand control commands, ObjRetarget employs a unified temporal scheduler to coordinate the execution of both arms and hands. The arm motions are carried out using the anthropomorphically retargeted commands produced by the arm-level optimization, ensuring physically plausible and smooth trajectories. Hand motions are executed according to a contact-aware strategy. When no contact is detected, defined as a hand–object interaction distance greater than the threshold $\delta$, the system follows the initialized hand retargeting commands. The threshold $\delta$ is set to 0.05cm, chosen based on the precision of the object tracking system to reliably distinguish actual contact from free motion. Once contact occurs, the hand control switches to the polyhedral-cluster-optimized commands to preserve local contact structures and manipulation semantics. After the contact is released, the system reverts to the initialized hand commands. This cycle repeats automatically for multiple contact events during an operation. By synchronizing arm and hand motions in this way, the scheduler ensures stable, natural, and generalizable execution of complex object-interaction tasks.

\section{Experiments}

\subsection{Experimental Setup}
To systematically evaluate the stability and generalization ability of ObjRetarget in real-world manipulation scenarios, we conduct experiments on a RealMan dual-arm robot platform across a range of everyday dexterous manipulation tasks.
The tasks cover grasping, placement, sequential pouring control, and articulated-object interaction, enabling evaluation under varying interaction complexity and motion constraints. Specifically, the tasks include: a) sequentially placing two cartons onto a tray, b) grasping a plush toy and placing it into a basket, c) placing a medicine box into an open drawer and closing it, d) transferring a mango from the table to a fruit basket, e) picking up a bottle, pouring water into a container, and returning it, and f) bimanual grasping of an apple and a lemon followed by synchronized placement.

The experiments are performed on the RealMan dual-arm robot, where each arm has 6 DoF and is equipped with a 6-DoF Inspire dexterous hand.
An Intel RealSense D435i depth camera is used for visual perception and experiment recording.
The robot executes motions using joint-space position control for continuous trajectory tracking, ensuring smooth, synchronized, and stable dual-arm execution.

For evaluation, each task is repeated 20 times.
In every trial, object poses are randomly initialized within the intersection of the camera field of view and the robot's reachable workspace, and experiments are conducted in cluttered tabletop environments containing distractor objects. This setup allows us to assess robustness under realistic perception uncertainty.

We compare ObjRetarget with OKAMI \cite{a32} and ORION \cite{a33}.
Since ORION was originally designed for parallel grippers, its generated end-effector trajectories cannot be directly applied to dexterous hands.
To ensure fair comparison, we adapt its control interface by converting the gripper trajectory into a palm trajectory and incorporating hand pose and joint constraints in the inverse kinematics solver, enabling stable multi-finger grasping and placement under identical task settings.

All experiments were conducted using the PyTorch deep learning framework on a system equipped with an NVIDIA RTX 4090 GPU and an Intel(R) Core(TM) i7-11700KF CPU.

\begin{table}[t]
\centering
\caption{Task Success Rate (\%)}
\label{tab:task_success}
\begin{tabular}{lccc}
\toprule
Task (/20f) & OKAMI\cite{a32} & ORION\cite{a33} & Ours \\
\midrule
Soft Place        & 13 & 12 & \textbf{17} \\
Pour Water        & 12 & 12 & \textbf{16} \\
Fruit Place       & 14 & 13 & \textbf{16} \\
Drawer Close      & 10 &  8 & \textbf{13} \\
Sequential Place  & 12 &  9 & \textbf{14} \\
Bimanual Place    & 13 &  8 & \textbf{15} \\
\midrule
Avg. Success Rate   & 61.6\% & 50.8\% & \textbf{75.8\%} \\
\bottomrule
\end{tabular}
\end{table}

\subsection{Comparative Experiments on Motion Retargeting}

\subsubsection{Multi-method comparison}
We conduct a systematic comparative evaluation of ObjRetarget, OKAMI, and ORION across the full task set to assess their execution stability and task completion capability in complex dexterous manipulation scenarios. The visualization of ObjRetarget executions is shown in Fig. \ref{fig4}. In all experiments, object poses are randomly initialized within the workspace, requiring each method to autonomously perform the tasks under varying spatial configurations and visual observations. This setup allows us to realistically evaluate robustness and generalization performance in cluttered and uncertain environments.

As shown in Table~\ref{tab:task_success}, ObjRetarget achieves the highest success rates across all tasks, with particularly clear gains in contact-sensitive and multi-stage scenarios such as Pour Water, Drawer Close, Sequential Place, and Bimanual Place. These results indicate that structured hand–object modeling improves contact stability, while unified arm-motion constraints enhance temporal consistency and coordination. OKAMI performs comparably on simpler placement tasks but degrades in operations requiring sustained contact or staged transitions. ORION yields the lowest overall performance, suggesting that without explicit motion and contact modeling, stable grasping and coordinated wrist behavior are difficult to maintain. 

\subsubsection{Cross-demonstrator generalization}
To further evaluate the system’s robustness to individual differences, we tested ObjRetarget under cross-demonstrator conditions. Three demonstrators with distinct hand sizes, motion habits, and manipulation styles performed the same set of tasks. Without modifying the model structure or parameters, the demonstrations from all individuals were mapped onto the same robot platform for execution.

\begin{figure}
\includegraphics[width=\linewidth]{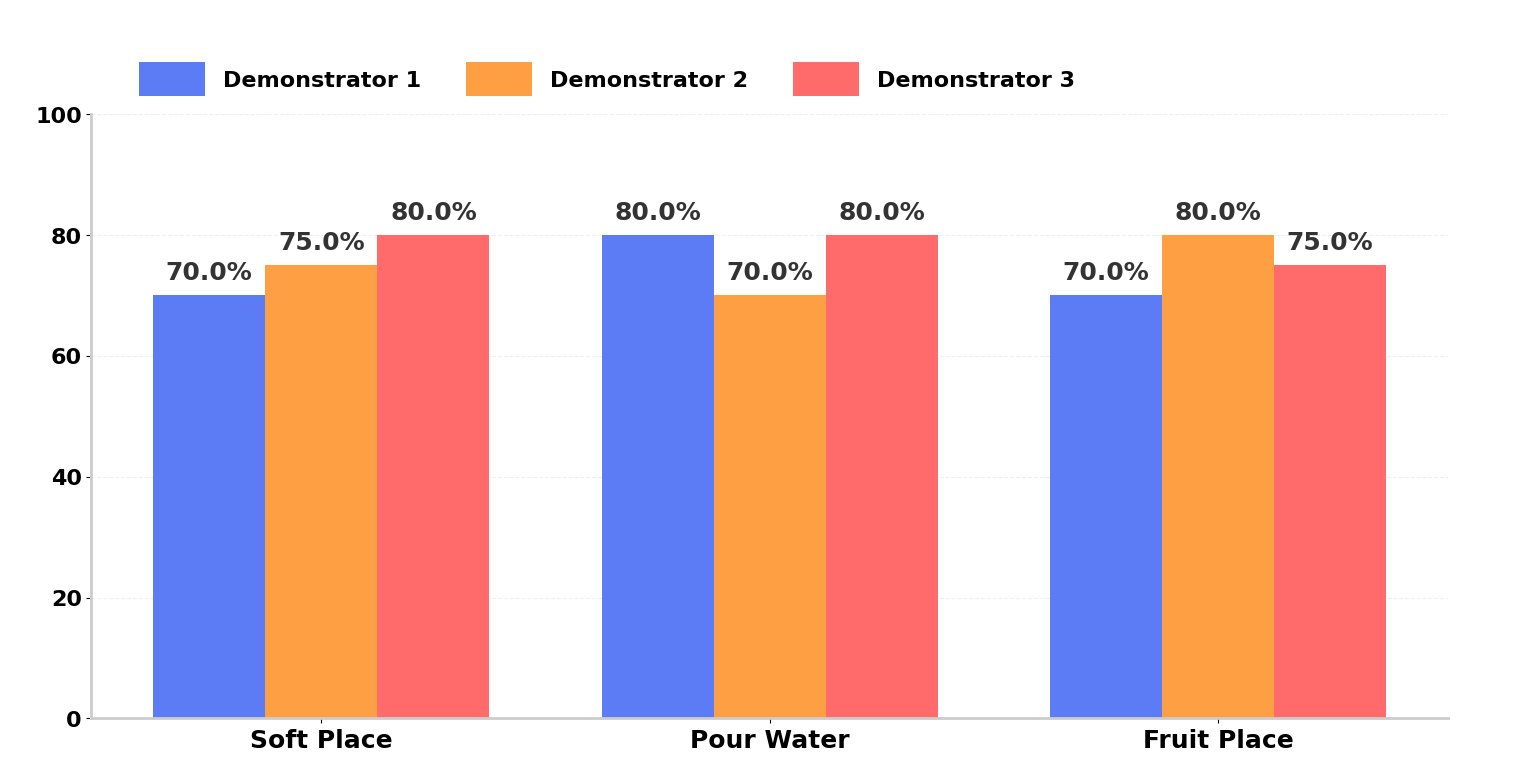}
\centering
\caption{Performance evaluation of ObjRetarget with videos from multiple demonstrators.}\label{fig5}
\end{figure}

As shown in Fig. \ref{fig5}, ObjRetarget achieves stable performance across demonstrations from different individuals. Although variations in manipulation style and hand morphology introduce slight differences in task outcomes, the success rates remain consistently high for all demonstrators across the three tasks. In particular, each task maintains comparable performance ranges, indicating that the system does not rely on a specific motion style to function correctly. 

\subsection{Ablation Study}
\subsubsection{Hand–object geometric consistency module}We assess the contribution of structured contact modeling to task stability and success by comparing three configurations: removing the human-demonstration-based initialization (W/o Init Retarget), further removing hand–object geometric consistency (W/o Hand Geometry), and the full ObjRetarget. Evaluation metrics include the object slip distance \cite{a34}, which measures the cumulative displacement of the object relative to the hand contact frame and thus reflects contact stability; geometric consistency \cite{a35}, computed from normalized edge-length errors and relative pose errors of local hand–object contacts, which indicates how well the contact structure is preserved; and task success, the percentage of trials in which the task is completed successfully.

\begin{table}[ht]
\centering
\caption{Ablation study of hand–object geometric consistency.}
\label{tab:ablation_hand_geometry}
\begin{tabular}{lccc}
\toprule
Method & Object Slip  & Geom.  & Task  \\
& Distance(m) $\downarrow$ & Consistency $\downarrow$ & Success(\%) $\uparrow$
\\
\midrule
W/o Init Retarget & 0.025 & 0.183 & 65.0 \\
W/o Hand Geom. & 0.045 & 0.275 & 53.3 \\
Ours & \textbf{0.012} & \textbf{0.081} &  \textbf{75.8}\\
\bottomrule
\end{tabular}
\end{table}

As shown in Table~\ref{tab:ablation_hand_geometry}, full ObjRetarget achieves the lowest geometric errors, minimal object sliding, and the highest success rates. Removing initialization slows convergence and reduces stability, while also removing geometric consistency causes significant slippage, contact failures, and posture collapse. 

\subsubsection{Ablation of Arm Initialization and Anthropomorphic Constraints}We evaluate the effects of the initialization-based retargeting and task-adaptive arm-plane regularization on motion accuracy and naturalness. Three configurations are compared: W/o Init Retarget (removing the initialization retargeting prior), W/o Arm-Plane Opt (removing arm-plane regularization), and full ObjRetarget. Performance is evaluated using four metrics: mean per-joint position error (MPJPE) for spatial accuracy \cite{a36}, end-effector quaternion distance (Quat) for orientation accuracy \cite{a37}, Fréchet distance for overall trajectory shape similarity \cite{a38}, and task success rate for execution reliability.

\begin{table}[h]
\centering
\caption{Ablation of Arm Initialization and Anthropomorphic Constraints}
\label{tab:1111}
\begin{tabular}{lcccc}
\toprule
{Method} & MPJPE& Quat & Fréchet & Task \\ & (m)$\downarrow$ & (rad)$\downarrow$ & (m)$\downarrow$ & Success (\%)$\uparrow$
\\
\midrule
W/o Init Retarget & 0.125 & 0.275 & 0.313 & 60.8 \\
W/o Arm-Plane Opt & 0.103 & 0.337 & 0.251 & 57.5 \\
Ours & \textbf{0.088} & \textbf{0.192} & \textbf{0.124} & \textbf{75.8} \\
\bottomrule
\end{tabular}
\end{table}

As shown in Table~\ref{tab:1111}, full ObjRetarget achieves the best results across all metrics, producing trajectories closely aligned with human demonstrations. Removing initialization increases MPJPE and Fréchet distance due to a larger search space, while removing the arm-plane regularization causes unnatural elbow motion and oscillations, reducing trajectory consistency. 

\section{Conclusion}
This paper presents ObjRetarget, a human-to-robot retargeting framework that decouples arm-level global motion from contact-sensitive hand manipulation by integrating polyhedral hand modeling and anthropomorphic arm constraints to achieve stable, natural, and generalizable dexterous manipulation. The framework is validated on the RMC-DA dual-arm robotic platform across diverse fine manipulation tasks, including grasping, pouring, object transfer, drawer closing, sequential placement, and bimanual coordination. Experimental results demonstrate that the proposed hand polyhedral cluster representation effectively preserves hand–object contact structures and manipulation semantics, while the reference-trajectory-guided arm optimization ensures accurate and anthropomorphically consistent global motion.

Despite these advantages, the current framework still has limitations: dynamic feedback in complex hand–object interactions is not yet fully exploited, and the method is presently limited to upper-limb manipulation. Future work will focus on integrating adaptive force feedback and extending ObjRetarget to whole-body robotic manipulation, aiming to enable more general and compliant dexterous manipulation systems.

\clearpage

\end{document}